\documentclass[11pt,a4paper]{article}
\usepackage[hyperref]{acl2020}
\usepackage{times}
\usepackage{latexsym}

\usepackage{microtype}

\aclfinalcopy 
\def\aclpaperid{554} 

\usepackage{xspace}
\usepackage{booktabs}
\usepackage{graphicx}
\usepackage{enumitem}
\setlist[description]{leftmargin=\parindent,labelindent=0pt}

\usepackage{tikz}
\usetikzlibrary{positioning}

\newcommand{\sref}[1]{Section~\ref{sec:#1}}
\newcommand{\srefplural}[1]{Sections~\ref{sec:#1}}
\newcommand{\srefshort}[1]{\ref{sec:#1}}
\newcommand{\tref}[1]{Table~\ref{tab:#1}}
\newcommand{\fref}[1]{Figure~\ref{fig:#1}}

\newcommand{\sofcCorpusName}{SOFC-Exp\xspace}
\newcommand{\sofcCorpusSize}{45\xspace} 

\newcommand{\expLabel}{Experiment\xspace}
\newcommand{\noExpLabel}{No-Experiment\xspace}

\newcommand{\matLabel}{\textsc{Material}\xspace}
\newcommand{\valLabel}{\textsc{Value}\xspace}
\newcommand{\devLabel}{\textsc{Device}\xspace}
\newcommand{\expLabelMention}{\textsc{Experiment}\xspace}

\newcommand{\sameExpLink}{\textit{same\_exp}\xspace}
\newcommand{\varExpLink}{\textit{exp\_variation}\xspace}

\newcommand{\slotInterlayerMaterial}{\textit{InterlayerMaterial}\xspace}
\newcommand{\slotResistance}{\textit{Resistance}\xspace}
\newcommand{\slotFuelUsed}{\textit{FuelUsed}\xspace}

\newcommand{\slotAnodeMaterial}{\textit{AnodeMaterial}\xspace}
\newcommand{\slotCurrentDensity}{\textit{CurrentDensity}\xspace}
\newcommand{\slotVoltage}{\textit{Voltage}\xspace}
\newcommand{\slotSupportMaterial}{\textit{SupportMaterial}\xspace}
\newcommand{\slotOpenCircuitVoltage}{\textit{OpenCircuitVoltage}\xspace}
\newcommand{\slotDegradationRate}{\textit{DegradationRate}\xspace}
\newcommand{\slotWorkingTemperature}{\textit{WorkingTemperature}\xspace}
\newcommand{\slotPowerDensity}{\textit{PowerDensity}\xspace}
\newcommand{\slotTimeOfOperation}{\textit{TimeOfOperation}\xspace}
\newcommand{\slotCathodeMaterial}{\textit{CathodeMaterial}\xspace}
\newcommand{\slotDevice}{\textit{Device}\xspace}
\newcommand{\slotElectrolyteMaterial}{\textit{ElectrolyteMaterial}\xspace}
\newcommand{\slotElectrolyte}{\textit{Electrolyte}\xspace}
\newcommand{\slotThickness}{\textit{Thickness}\xspace}
\newcommand{\slotConductivity}{\textit{Conductivity}\xspace}

\usepackage{adjustbox}
\usepackage{array}

\newcolumntype{R}[2]{%
	>{\adjustbox{angle=#1,lap=\width-(#2)}\bgroup}%
	l%
	<{\egroup}%
}
\newcommand*\rot{\multicolumn{1}{R{60}{1em}}}

\usepackage{siunitx} 
\newcommand{\pd}{\si{\watt\per\centi\metre\squared}\xspace} 

\title{The \sofcCorpusName Corpus and Neural Approaches\\ to Information Extraction in the Materials Science Domain}

\author{\parbox{\linewidth}{\centering Annemarie Friedrich$^{1}$ \hspace*{5mm} Heike Adel$^{1}$ \hspace*{5mm} Federico Tomazic$^2$\hspace*{5mm} Johannes Hingerl$^1$ \\  Renou Benteau$^{1}$ \hspace*{1cm} Anika Maruscyk$^2$ \hspace*{1cm} Lukas Lange$^{1}$}
\vspace*{2mm}\\
  $^1$Bosch Center for Artificial Intelligence, Renningen, Germany\\
  $^2$Corporate Research, Robert Bosch GmbH, Renningen, Germany\\
  \texttt{firstname.lastname@de.bosch.com} \\}

\date{}

\begin{document}
\maketitle

\begin{abstract}
  This paper presents a new challenging information extraction task in the domain of materials science.
We develop an annotation scheme for marking information on experiments related to solid oxide fuel cells in scientific publications, such as involved materials and measurement conditions.
With this paper, we publish our annotation guidelines, as well as our \sofcCorpusName corpus consisting of \sofcCorpusSize open-access scholarly articles annotated by domain experts.
A corpus and an inter-annotator agreement study demonstrate the complexity of the suggested named entity recognition and slot filling tasks as well as high annotation quality.
We also present strong neural-network based models for a variety of tasks that can be addressed on the basis of our new data set.
On all tasks, using BERT embeddings leads to large performance gains, but with increasing task complexity, adding a recurrent neural network on top seems beneficial.
Our models will serve as competitive baselines in future work, and analysis of their performance highlights difficult cases when modeling the data and suggests promising research directions.
\end{abstract}

\section{Introduction}
\label{sec:intro}
The design of new experiments in scientific domains heavily depends on domain knowledge as well as on previous studies and their findings.
However, the amount of publications available is typically very large, making it hard or even impossible to keep track of all experiments conducted for a particular research question.
Since scientific experiments are often time-consuming and expensive, effective knowledge base population methods for finding promising settings based on the published research would be of great value \citep[e.g.,][]{auer2018towards,manica2019information,stroetgen2019towards,mrdjenovich2020propnet}.
While such real-life information extraction tasks have received considerable attention in the biomedical domain \citep[e.g.,][]{cohen2017bionlp,demner2018bionlp,demner2019bionlp}, there has been little work in other domains \citep{nastase2019extracting}, including materials science \citep[with the notable exception of the work by][]{mysore2017automatically,mysore2019materials}.

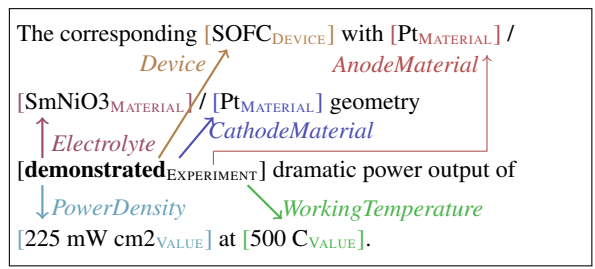
\begin{figure}[t]
	\small
	\noindent\fbox{%
		\parbox{0.46\textwidth}{%
			\hspace*{-2mm}	
			\begin{tikzpicture}[scale=0.9]
			\node [anchor=west] (a) at (0,0) {The corresponding \textcolor{orange!40!gray}{[}SOFC\textcolor{orange!40!gray}{$_{\devLabel}$]} with \textcolor{red!40!gray}{[}Pt\textcolor{red!40!gray}{$_{\matLabel}$]} /};
			\node [anchor=west] (b) at (0,-1) {\textcolor{purple!40!gray}{[}SmNiO3\textcolor{purple!40!gray}{$_{\matLabel}$]} / \textcolor{blue!40!gray}{[}Pt\textcolor{blue!40!gray}{$_{\matLabel}$]} geometry };
			\node [anchor=west] (c) at (0,-2) {[\textbf{demonstrated}$_{\expLabelMention}$] dramatic power output of};
			\node [anchor=west] (d) at (0,-3) {\textcolor{cyan!40!gray}{[}225 mW cm−2\textcolor{cyan!40!gray}{$_{\valLabel}$]} at \textcolor{green!40!gray}{[}500 °C\textcolor{green!40!gray}{$_{\valLabel}$]}.};
			
			\draw[thick,->,color=green!40!gray] (3.5,-2.2) -- (4,-2.7) node[midway,right,xshift=3pt,yshift=-3pt] {\slotWorkingTemperature};
			\draw[thick,->,color=cyan!40!gray] (.5,-2.2) -- (.5,-2.7) node[midway,right,yshift=-3pt] {\slotPowerDensity};
			\draw[thick,->,color=purple!40!gray] (.5,-1.8) -- (.5,-1.2) node[midway,right,yshift=-3pt] {\slotElectrolyte};
			\draw[thick,->,color=blue!40!gray] (2.5,-1.8) -- (3,-1.2) node[midway,right,yshift=3pt,xshift=1pt] {\slotCathodeMaterial};
			\draw[thick,->,color=orange!40!gray] (2.2,-1.8) -- (3.2,-.2) node[near end,left,xshift=2pt,yshift=5pt] {\slotDevice};
			\draw [->,color=red!40!gray] 
			(3,-1.7) -- +(4,0) |- node[near end,left,yshift=-3pt] {\slotAnodeMaterial} (7,-.3);
			\draw [-,color=red!40!gray]  (3,-1.9) -- (3,-1.7);  
			\end{tikzpicture}
		}
	}
	\caption{Sentence describing a fuel-cell related experiment, annotated with Experiment frame information.}
	\label{fig:annotation_example}
\end{figure}

In this paper, we introduce a new information extraction use case from the materials science domain and propose a series of new challenging information extraction tasks.
We target publications about solid oxide fuel cells (SOFCs) in which the interdependence between chosen materials, measurement conditions and performance is complex (see \fref{annotation_example}).
For making progress within natural language processing (NLP), the genre-domain combination presents interesting challenges and characteristics, e.g., domain-specific tokens such as material names and chemical formulas.

We provide a new corpus of open-access scientific publications annotated with semantic frame information on experiments mentioned in the text.
The annotation scheme has been developed jointly with materials science domain experts, who subsequently carried out the high-quality annotation.
We define an ``Experiment''-frame and annotate sentences that evoke this frame with a set of 16 possible slots, including among others \slotAnodeMaterial, \slotFuelUsed and \slotWorkingTemperature, reflecting the role the referent of a mention plays in an experiment.
Frame information is annotated on top of the text as graphs rooted in the experiment-evoking element (see \fref{annotation_example}).
In addition, slot-filling phrases are assigned one of the types \matLabel, \valLabel, and \devLabel.

The task of finding experiment-specific information can be modeled as a retrieval task (i.e., finding relevant information in documents) and at the same time as a semantic-role-labeling task (i.e., identifying the slot fillers).
We identify three sub-tasks:
(1) identifying sentences describing relevant experiments, (2) identifying mentions of materials, values, and devices, and (3) recognizing mentions of slots and their values related to these experiments.
We propose and compare several machine learning methods for the different sub-tasks, including bidirectional long-short term memory (BiLSTM) networks and BERT-based models.
In our results, BERT-based models show superior performance. However, with increasing complexity of the task, it is beneficial to combine the two approaches.

With the aim of fostering research on challenging information extraction tasks in the scientific domain, we target the domain of SOFC-related experiments as a starting point.
Our findings based on this sample use case are transferable to similar experimental domains, which we illustrate by applying our best model configurations to a previously existing related corpus \citep{mysore2019materials}, achieving state-of-the-art results.

We sum up our contributions as follows:
\begin{itemize}
	\setlength\itemsep{5pt}
	\item We develop an annotation scheme for marking information on materials-science experiments on scientific publications (\sref{annotationScheme}).
	\item We provide a new corpus of \sofcCorpusSize materials-science publications in the research area of SOFCs, manually annotated by domain experts for information on experimental settings and results (\sref{data}). Our corpus is publicly available.\footnote{Resources related to this paper can be found at:\\ \url{https://github.com/boschresearch/sofc-exp_textmining_resources}} Our inter-annotator agreement study provides evidence for high annotation quality (\sref{agreement}).
	\item We identify three sub-tasks of extracting experiment information and provide competitive baselines with state-of-the-art neural network approaches for them (\srefplural{data}, \srefshort{modeling}, \srefshort{experiments}).
	\item We show the applicability of our findings to modeling the annotations of another materials-science corpus \citep[][\sref{experiments}]{mysore2019materials}.
\end{itemize}
\section{Related work}
\label{sec:relwork}

\textbf{Information extraction for scientific publications.}
Recently, several studies addressed information extraction and knowledge base construction in the scientific domain \cite{augenstein2017semeval,luan2018multitask,jiang2019role,buscaldi2019mining}.
We also aim at knowledge base construction but target publications about materials science experiments, a domain understudied in NLP to date.

\textbf{Information extraction for materials science.}
The work closest to ours is the one of \citet{mysore2019materials} who annotate a corpus of 230 paragraphs describing synthesis procedures with operations and their arguments, e.g., ``The resulting [solid products$_{Material}$] were ... 
[dried$_{Operation}$] at [120$_{Number}$][\si{celsius}$_{ConditionUnit}$] for [8$_{Number}$] [h$_{ConditionUnit}$].''
Operation-evoking elements (``dried'') are connected to their arguments via links, and with each other to indicate temporal sequence, thus resulting in graph structures similar to ours.
Their annotation scheme comprises 21 entity types and 14 relation types such as \textit{Participant-material}, \textit{Apparatus-of} and \textit{Descriptor-of}.
\citet{kononova2019text} also retrieve synthesis procedures and extract recipes, though with a coarser-grained label set, focusing on different synthesis operation types.
\citet{weston2019named} create a dataset for named entity recognition on abstracts of materials science publications.
In contrast to our work, their label set (e.g., \textit{Material}, \textit{Application}, \textit{Property}) is targeted to document indexing rather than information extraction.
A notable difference to our work is that we perform full-text annotation while the aforementioned approaches annotate a pre-selected set of paragraphs \citep[see also][]{kim2017machine}.

\citet{mysore2017automatically} apply the generative model of \citet{kiddon2015mise} to induce \textit{action graphs} for synthesis procedures of materials from text.
In \sref{synthesisExp}, we implement a similar entity extraction system and also apply our algorithms to the dataset of \citet{mysore2019materials}.
\citet{mat2vec} train word2vec \cite{word2vec} embeddings on materials science publications and show that they can be used for recommending materials for functional applications.
Other works adapt the BERT model to clinical and biomedical domains \citep{alsentzer2019publicly,sun2019transfer}, or generally to scientific text \citep{beltagy2019scibert}.

\textbf{Neural entity tagging and slot filling.}
The neural-network based models we use for entity tagging and slot filling bear similarity to state-of-the-art models for named entity recognition \citep[e.g.,][]{huang2015,lample-etal-2016-neural,panchendrarajan-amaresan-2018-bidirectional,lange2019nlnde}.
Other related work exists in the area of semantic role labeling \citep[e.g.,][]{roth2015context,kshirsagar2015frame,hartmann2017domain,adel-etal-2018-dere,swayamdipta2018syntactic}.
\section{Annotation Scheme}
\label{sec:annotationScheme}
In this section, we describe our annotation scheme and guidelines for marking information on SOFC-related experiments in scientific publications.

\subsection{Experiment-Describing Sentences}
We treat the annotation task as identifying instances of a semantic frame \cite{fillmore1976frame} that represents SOFC-related experiments.
We include (1) cases that introduce novel content; (2) descriptions of specific previous work; (3) general knowledge that one could find in a textbook or survey; and also (4) suggestions for future work.

We assume that a frame is introduced to the discourse by words that \textit{evoke} the frame.
While we allow any part-of-speech for such frame-evoking elements, in practice, our annotators marked almost only verbs, such as ``test,'' ``perform,'' and ``report'' with the type \expLabelMention.
In the remainder of this paper, we treat all sentences containing at least one such annotation as experiment-describing.

\subsection{Entity Mention Types}

In a second annotation layer, annotators mark spans with one of the following entity types.
The annotations are marked only on experiment-describing sentences as well as several additional sentences selected by the annotator.
\begin{description}
	\setlength\itemsep{1pt}
	\item[\matLabel.] We use the type \matLabel to annotate text spans referring to materials or elements. They may be specified by a particular composition formula (e.g., ``La$_{0.75}$Sr$_{0.25}$Cr$_{0.5}$Mn$_{0.5}$O$_3$'') or just by a mention of the general class of materials, such as ``oxides'' or ``hydrocarbons.''\footnote{If the material is referenced by a common noun or by a pronoun and a more specific mention occurs earlier in the text, we indicate this coreference with the aim of facilitating oracle information extraction experiments in future work.}

	\item[\valLabel.] We annotate numerical values and their respective units with the type \valLabel. 

	In addition, we include specifications like ``more than'' or ``between'' in the annotation span (e.g., ``above 750 \si{\celsius},''  ``1.0 \pd'').
	\item[\devLabel.] This label is used to mark mentions of the type of device used in the fuel cell experiment (e.g., ``IT-SOFC'').
\end{description}

\subsection{Experiment Slot Types}
The above two steps of recognizing relevant sentences and marking coarse-grained entity types are in general applicable to a wide range of experiment types within the materials science domain.
We now define a set of slot types particular to experiments on SOFCs.
During annotation, we mark these slot types as links between the experiment-evoking phrase and the respective slot filler (entity mention), see \fref{annotation_example}.
As a result, experiment frames are represented by graphs rooted in the node corresponding to the frame-evoking element.

Our annotation scheme comprises 16 slot types relevant for SOFC experiments.
Here we explain a few of these types for illustration.
A full list of these slot types can be found in Supplementary Material \tref{slot-agreement}; detailed explanations are given in the annotation guidelines published along with our corpus.

\begin{description}
	\setlength\itemsep{1pt}
	\item[\slotAnodeMaterial, \slotCathodeMaterial:] These slots are used to mark the fuel cell's anode and cathode, respectively.
	Both are entity mentions of type \matLabel.
	In some cases, simple surface information indicates that a material fulfills such a role.
	Other cases require specific domain knowledge and close attention to the context.
	
	\item[\slotFuelUsed:] This slot type indicates the chemical composition or the class of a fuel or the oxidant species (indicated as a \matLabel).
	
	\item[\slotPowerDensity, \slotResistance, \slotWorkingTemperature:] These slots are generally filled by mentions of type \valLabel, i.e., a numerical value plus a unit.
	Our annotation guidelines give examples for relevant units and describe special cases.
	This enables any materials scientist, even if he/she is not an expert on SOFCs, to easily understand and apply our annotation guidelines.
\end{description}

\paragraph{Difficult cases.}
We also found sentences that include enumerations of experimental settings such as in the following example:
``It can be seen that the electrode polarization resistances in air are \SI{0.027}{\ohm}cm$^2$, \SI{0.11}{\ohm}cm$^2$, and \SI{0.88}{\ohm}cm$^2$ at \SI{800}{\celsius}, \SI{700}{\celsius} and \SI{600}{\celsius}, respectively.''\footnote{See [PMC4673446].}
We decided to simply link all slot fillers (the various resistance and temperature values) to the same frame-evoking element, leaving disentangling and grouping of this set of parameters to future work.

\subsection{Links between Experiments}
We instruct our annotators to always link slot fillers to the syntactically closest \expLabelMention mention.
If the description of an experiment spans more than one clause, we link the two relevant \expLabelMention{}s using the relation \sameExpLink.
We use \varExpLink to link experiments done on the same cell, but with slightly different operating conditions.
The link type \varExpLink can also relate two frame-evoking elements that refer to two measurements performed on different materials/cells, but in the same experimental conditions.
In this case, the frame-evoking elements usually convey an idea of comparison, e.g., ``increase'' or ``reach from ... to.''


\section{Corpus Statistics and Task Definitions}
\label{sec:data}

In this section, we describe our new corpus and propose a set of information extraction tasks that can be trained and evaluated using this dataset.

\paragraph{\sofcCorpusName Corpus.}
Our corpus consists of \sofcCorpusSize open-access scientific publications about SOFCs and related research, annotated by domain experts.
For manual annotation, we use the InCeption annotation tool \citep{klie2018inception}.
\tref{corpusStats} shows the key statistics for our corpus.
Sentence segmentation was performed automatically.\footnote{InCeption uses Java's built-in sentence segmentation algorithm with US locale.}
As a preparation for experimenting with the data, we manually remove all sentences belonging to the Acknowledgment and References sections.
We propose the experimental setting of using the training data in a 5-fold cross validation setting for development and tuning, and finally applying the model(s) to the independent test set.

\begin{table}[t]
	\centering
	\footnotesize
	\begin{tabular}{lrr}
		\toprule
		& \textbf{train} & \textbf{test}\\
		\midrule
		documents & 34 & 11\\
		sentences & 7,630 & 1,836\\
		avg. token/sentence & 29.4 & 35.0\\
		\midrule
		experiment-describing sentences & 703 & 173\\
		\hspace{3mm} in \% & 9.2 & 9.4\\
		\midrule
		sentences with entity mention&\\
		annotations & 853 & 210\\
		\midrule
		entity mention annotations & 4,037 & 1058\\
		\hspace{3mm} \matLabel & 1,530 & 329\\
		\hspace{3mm} \valLabel & 1,177 & 370\\
		\hspace{3mm} \devLabel & 468 & 130\\
		\hspace{3mm} \expLabelMention & 862 & 229\\
		\bottomrule
	\end{tabular}
\caption{\textbf{\sofcCorpusName corpus annotation statistics.}}
\label{tab:corpusStats}
\end{table}

\paragraph{Task definitions.}
Our rich graph-based annotation scheme allows for a number of information extraction tasks.
In the scope of this paper, we address the following steps of (1) identifying sentences that describe SOFC-related experiments, (2) recognizing and typing relevant named entities, and (3) extracting slot fillers from these sentences.
The originally annotated graph structures would also allow for modeling as relations or dependency structures.
We leave this to future work.

The setup of our tasks is based on the assumption that in most cases, one sentence describes a single experiment.
The validity of this assumption is supported by the observation that in almost all sentences containing more than one \expLabelMention, experiment-evoking verbs actually describe variations of the same experiment.
(For details on our analysis of links between experiments, see Supplementary Material \sref{dataAnalysis}.)
In our automatic modeling, we treat slot types as entity-types-in-context, which is a valid approximation for information extraction purposes.
We leave the tasks of deciding whether two experiments are the same (\sameExpLink) or whether they constitute a variation (\varExpLink) to future work.
While our dataset provides a good starting point, tackling these tasks will likely require collecting additional data.


\section{Inter-annotator Agreement Study}
\label{sec:agreement}
We here present the results of our inter-annotator agreement study, which we perform in order to estimate the degree of reproducibility of our corpus and to put automatic modeling performance into perspective.
Six documents (973 sentences) have been annotated independently both by our primary annotator, a graduate student of materials science, and a second annotator, who holds a Ph.D. in physics and is active in the field of materials science.
The label distribution in this subset is similar to the one of our overall corpus, with each annotator choosing \expLabelMention about 11.8\% of the time.\looseness=-1

\paragraph{Identification of experiment-describing sentences.}
Agreement on our first task, judging whether a sentence contains relevant experimental information, is 0.75 in terms of Cohen's $\kappa$ \citep{cohen1968weighted}, indicating substantial agreement according to \citet{landis1977measurement}.
The observed agreement, corresponding to accuracy, is 94.9\%; expected agreement amounts to 79.2\%.
\tref{agreement-ir-stats} shows precision, recall and F1 for the doubly-annotated subset, treating one annotator as the gold standard and the other one's labels as predicted.
Our primary annotator identifies 119 out of 973 sentences as experiment-describing, our secondary annotator 111 sentences, with an overlap of 90 sentences.
These statistics are helpful to gain further intuition of how well a human can reproduce another annotator's labels and can also be considered an upper bound for system performance.

\begin{table}[t]
	\centering
	\footnotesize
	\begin{tabular}{lcccc}
		\toprule
		& \multicolumn{1}{c}{\textbf{P}} & \multicolumn{1}{c}{\textbf{R}} & \multicolumn{1}{c}{\textbf{F1}} & \multicolumn{1}{c}{\textbf{count}}\\ \midrule
		\expLabel & 81.1 & 75.6 & 78.3 & 119\\
		\noExpLabel & 96.6 & 97.5 & 97.1 & 854\\
		\bottomrule
	\end{tabular}
	\caption{\textbf{Inter-annotator agreement study.} Precision, recall and F1 for the subset of doubly-annotated documents. \textbf{count} refers to the number of mentions labeled with the respective type by our primary annotator.}
	\label{tab:agreement-ir-stats}
\end{table}

\paragraph{Entity mention detection and type assignment.}
As mentioned above, relevant entity mentions and their types are only annotated for sentences containing experiment information and neighboring sentences.
Therefore, we here compute agreement on the detection of entity mention and type assignment on the subset of 90 sentences that both annotators considered as containing experimental information.
We again look at precision and recall of the annotators versus each other, see \tref{agreement-ir-stats-mentions}.
The high precision indicates that our secondary annotator marks essentially the same mentions as our primary annotator, but recall suggests a few missing cases.
The difference in marking \expLabelMention can be explained by the fact that the primary annotator sometimes marks several verbs per sentence as experiment-evoking elements, connecting them with \sameExpLink or \varExpLink, while the secondary annotator links the mentions of relevant slots to the first experiment-evoking element (see also Supplementary Material \sref{dataAnalysis}).
Overall, the high agreement between domain expert annotators indicates high data quality.

\begin{table}[t]
	\centering
	\footnotesize
	\begin{tabular}{lrrrr}
		\toprule
		& \multicolumn{1}{c}{\textbf{P}} & \multicolumn{1}{c}{\textbf{R}} & \multicolumn{1}{c}{\textbf{F1}} & \multicolumn{1}{c}{\textbf{count}}\\ \midrule
		\expLabelMention & 100.0 & 89.3 & 94.3 & 112\\
		\matLabel & 100.0 & 92.1 & 95.9 & 190\\
		\valLabel & 100.0 & 91.5 & 95.5 & 211\\
		\devLabel & 96.3 & 98.7 & 97.5 & 78\\
		\bottomrule
	\end{tabular}
	\caption{\textbf{Inter-annotator agreement study.} Precision, recall and F1 for labeling entity types.
		\textbf{count} refers to the number of mentions labeled with the respective type by our primary annotator.}
	\label{tab:agreement-ir-stats-mentions}
\end{table}

\paragraph{Identifying experiment slot fillers.}
We compute agreement on the task of identifying the slots of an experiment frame filled by the mentions in a sentence on the subset of sentences that both annotators marked as experiment-describing.
Slot fillers are the dependents of the respective edges starting at the experiment-evoking element.
\tref{slot-agreementShort} shows F1 scores for the most frequent ones among those categories.
See Supplementary Material \sref{suppl:agreement} for all slot types.
Overall, our agreement study provides support for the high quality of our annotation scheme and validates the annotated dataset.

\begin{table}[t]
	\centering
	\footnotesize
	\begin{tabular}{lrrr}
		\toprule
		& & \textbf{IAA} & \textbf{train}\\
		& \multicolumn{1}{c}{\textbf{F1}} & \textbf{count} & \textbf{count}\\ \midrule
		\slotAnodeMaterial & 72.0 & 13 & 280\\
		\slotCathodeMaterial & 86.7 & 44 & 259\\
		\slotDevice & 95.0 & 71 & 381\\
		\slotElectrolyteMaterial & 85.7 & 48 & 219\\
		\slotFuelUsed & 85.7 & 11 & 159\\
		\slotInterlayerMaterial & 71.8 & 25 & 51\\
		\slotOpenCircuitVoltage & 90.0 & 10 & 44\\
		\slotPowerDensity & 92.0 & 47 & 175\\
		\slotResistance & 100.0 & 26 & 136\\
		\slotThickness & 92.6 & 27 & 83\\
		\slotWorkingTemperature & 96.5 & 73 & 414\\
		\bottomrule
	\end{tabular}
	\caption{\textbf{Inter-annotator agreement study.} F1 was computed for the two annotators vs. each other on the set of \textbf{experiment slots}; \textbf{IAA count} refers to the number of mentions labeled with the respective type by our primary annotator in the inter-annotator agreement study (IAA).}
	\label{tab:slot-agreementShort}
\end{table}

\section{Modeling}
\label{sec:modeling}
In this section, we describe a set of neural-network based model architectures for tackling the various information extraction tasks described in \sref{data}.

\paragraph{Experiment detection.}
The task of experiment detection can be modeled as a binary sentence classification problem.
It can also be conceived as a retrieval task, selecting sentences as candidates for experiment frame extraction.
We implement a bidirectional long short-term memory (\textbf{BiLSTM}) model with attention for the task of experiment sentence detection.
Each input token is represented by a concatenation of several pretrained word embeddings, each of which is fine-tuned during training.
We use the Google News word2vec embeddings \cite{word2vec}, domain-specific word2vec embeddings \citep[mat2vec,][see also \sref{relwork}]{mat2vec}, subword embeddings based on byte-pair encoding \citep[bpe,][]{bpemb/heinzerling18}, BERT \cite{devlin2019bert}, and SciBERT \citep{beltagy2019scibert} embeddings.
For BERT and SciBERT, we take the embeddings of the first word piece as token representation.
The embeddings are fed into a BiLSTM model followed by an attention layer that computes a vector for the whole sentence.
Finally, a softmax layer decides whether the sentence contains an experiment.

In addition, we fine-tune the original (uncased) \textbf{BERT} \cite{devlin2019bert} as well as \textbf{SciBERT} \citep{beltagy2019scibert} models on our dataset.
Sci-BERT was trained on a large corpus of scientific text.
We use the implementation of the BERT sentence classifier by \citet{Wolf2019HuggingFacesTS} that uses the CLS token of BERT as input to the classification layer.\footnote{\url{https://github.com/huggingface/transformers}}

Finally, we compare the neural network models with traditional classification models,
namely a support vector machine (\textbf{SVM}) and a \textbf{logistic regression} classifier.
For both models, we use the following set of input features: bag-of-words vectors indicating which 1- to 4-grams and part-of-speech tags occur in the sentence.\footnote{We use sklearn, \url{https://scikit-learn.org}.}

\paragraph{Entity mention extraction.}
\label{sec:model:entitymention}
For entity and concept extraction, we use a sequence-tagging approach similar to \citep{huang2015,lample-etal-2016-neural}, namely a \textbf{BiLSTM model}.
We use the same input representation (stacked embeddings) as above, which are fed into a BiLSTM.
The subsequent conditional random field \citep[CRF,][]{lafferty2001CRF} output layer extracts the most probable label sequence.
To cope with multi-token entities, we convert the labels into BIO format.

We also fine-tune the original \textbf{BERT} and \textbf{SciBERT} sequence tagging models on this task.
Since we use BIO labels, we extend it with a CRF output layer to enable it to correctly label multi-token mentions and to enable it to learn transition scores between labels.
As a non-neural baseline, we train a \textbf{CRF} model using the token, its lemma, part-of-speech tag and mat2vec embedding as features.\footnote{We use sklearn-pycrfsuite, \url{https://pypi.org/project/sklearn-pycrfsuite}.}

\paragraph{Slot filling.}
As described in \sref{data}, we approach the slot filler extraction task as fine-grained entity-typing-in-context, assuming that each sentence represents a single experiment frame.
We use the same sequence tagging architectures as above for tagging the tokens of each experiment-describing sentence with the set of slot types (see \tref{slot-agreement}).
Future work may contrast this sequence tagging baseline with graph-induction based frame extraction.

\section{Experiments}
\label{sec:experiments}

In this section, we present the experimental results for detecting experiment-describing sentences, entity mention extraction and experiment slot identification.
For tokenization, we employ ChemDataExtractor,\footnote{\url{http://chemdataextractor.org}} which is optimized for dealing with chemical formulas and unit mentions.

We tune our models in a 5-fold cross-validation setting.
We also report the mean and standard deviation across those folds as development results.
For the test set, we report the macro-average of the scores obtained when applying each of the five models to the test set.
To put model performance in relation to human agreement, we report the corresponding statistics obtained from our inter-annotator agreement study (\sref{agreement}).
Note that these numbers are based on a subset of the data and are hence not directly comparable.

\paragraph{Hyperparameters and training.}
The BiLSTM models are trained with the Adam optimizer \cite{adam} with a learning rate of 1e-3.
For fine-tuning the original BERT models, we follow the configuration published by \newcite{Wolf2019HuggingFacesTS} and use AdamW \citep{loshchilov2019decoupled} as optimizer and a learning rate of 4e-7 for sentence classification and 1e-5 for sequence tagging.
When adding BERT tokens to the BiLSTM, we also use the AdamW optimizer for the whole model and learning rates of 4e-7 or 1e-5 for the BERT part and 1e-3 for the remainder.
For regularization, we employ early stopping on the development set.
We use a stacked BiLSTM with two hidden layers and 500 hidden units for all tasks with the exception of the experiment sentence detection task, where we found one BiLSTM layer to work best.
The attention layer of the sentence detection model has a hidden size of 100.

\begin{table}[t]
	\centering
	\footnotesize
	\begin{tabular}{lr|rrr}
		\toprule
		& \multicolumn{1}{c|}{\textbf{dev}}& \multicolumn{3}{c}{\textbf{test}}\\
		Model & F1 & P & R & F1\\
		\midrule
		RBF SVM &  54.2\tiny{+/-3.7} &    64.6 &    54.9 &    59.4\\
		Logistic Regression & 53.0\tiny{+/-4.2} &    \textbf{68.2} &    50.9 &    58.3\\
		\midrule
BiLSTM mat2vec &    49.9\tiny{+/-3.1} &    49.6 &    69.4 &    57.8\\
BiLSTM word2vec &    52.3\tiny{+/-4.6} &    51.1 &    65.3 &    57.4\\
\hspace*{2mm}+ mat2vec &    55.9\tiny{+/-4.2} &    52.0 &    59.0 &    55.3\\
\hspace*{4mm}+ bpe &    58.6\tiny{+/-3.0} &    58.9 &    64.7 &    61.7\\
\hspace*{6mm}+ BERT-base &    66.8\tiny{+/-4.9} &    60.2 &    71.7 &    65.4\\
\hspace*{6mm}+ SciBERT & 67.9\tiny{+/-4.0} &    58.6 &    74.6 &    65.6\\
BiLSTM BERT-base &   64.7\tiny{+/-4.6} &    63.7 &    69.9 &    66.7\\
BiLSTM SciBERT & \textbf{68.1}\tiny{+/-3.7} &    60.2 &    73.4 &    66.1\\
		\midrule
		BERT-base & 66.0\tiny{+/-4.6} &    58.6 &    71.1 &    64.2\\
		SciBERT & 67.9\tiny{+/-4.0} & 60.8 & 74.6 & 67.0\\
		BERT-large & 64.3\tiny{+/-4.3} & 63.1 & \textbf{75.1} & \textbf{68.6}\\
		\midrule
		\textit{humans} & \textit{78.3} & \textit{81.1} & \textit{75.6} & \textit{78.3}\\
		\bottomrule
	\end{tabular}
	\caption{\textbf{Experiments: identifying experiment-describing sentences.} P, R and F1 for experiment-describing sentences. With the exception of SVM, we downsample the non-experiment-describing sentences of the training set by 0.3.}
	\label{tab:results-experiment-detection}
\end{table}

\paragraph{Experiment sentence detection.}
\tref{results-experiment-detection} shows our results on the detection of experiment-describing sentences.
The neural models with byte-pair encoding embeddings or BERT clearly outperform the SVM and logistic regression models. 
Within the neural models, BERT and SciBERT add the most value, both when using their embeddings as another input to the BiLSTM and when fine-tuning the original BERT models.
Note that even the general-domain BERT is strong enough to cope with non-standard domains.
Nevertheless, models based on SciBERT outperform BERT-based models, indicating that in-domain information is indeed beneficial.
For performance reasons, we use BERT-base in our experiments, but for the sake of completeness, we also run BERT-large for the task of detecting experiment sentences.
Because it did not outperform BERT-base in our cross-validation based development setting, we did not further experiment with BERT-large.
However, we found that it resulted in the best F1-score achieved on our test set.
In general, SciBERT-based models provide very good performance and seem most robust across dev and test sets.
Overall, achieving F1-scores around 67.0-68.6, such a retrieval model may already be useful in production.
However, there certainly is room for improvement.
\paragraph{Entity mention extraction.}

\tref{results-entity-mention-extraction} provides our results on entity mention detection and typing.
Models are trained and results are reported on the subset of sentences marked as experiment-describing in the gold standard, amounting to 4,590 entity mentions in total.\footnote{The \sofcCorpusName gold standard marks all entity mentions that correspond to one of the four relevant types occurring in these sentences, regardless of whether the mention fills a slot in an experiment or not.}
The CRF baseline achieves comparable or better results than the Bi-LSTM with word2vec and/or mat2vec embeddings.
However, adding subword-based embeddings (bpe and/or BERT) significantly increases performance of the BiLSTM, indicating that there are many rare words.
Again, the best results are obtained when using BERT or SciBERT embeddings or when using the original SciBERT model.
It is relatively easy for all model variants to recognize \valLabel as these mentions usually consist of a number and unit which the model can easily memorize.
Recognizing the types \matLabel and \devLabel, in contrast, is harder and may profit from using gazetteer-based extensions.

\begin{table}[t]
	\centering
        \footnotesize
		\setlength{\tabcolsep}{7pt}
	\begin{tabular}{lp{0.45cm}p{0.45cm}p{0.45cm}p{0.45cm}p{0.5cm}}
		\toprule
\textbf{Model} & \textsc{Exp.} & \textsc{Mat.} & \textsc{Val.} &
		\textsc{Dev.} & \textbf{avg.}\\
		\midrule
CRF & 61.4 & 42.3 & 73.6 & 64.1 & 60.3\\
\midrule
BiLSTM mat2vec &    47.1 &    52.4 &    60.9 &    46.1 &    51.6\\
BiLSTM word2vec &   55.8 &    58.6 &    59.1 &    51.7 &    56.3\\
\hspace*{2mm}+mat2vec &    57.9 &    75.2 &    64.3 &    61.5 &    64.7\\
\hspace*{4mm}+bpe &     63.3 &    81.6 &    68.0 &    68.1 &    70.2\\
\hspace*{6mm}+BERT-base &    76.0 &    88.1 &    72.9 &    81.5 &   79.7\\
\hspace*{6mm}+SciBERT & 76.9 &    89.8 &    74.1 &    85.2 &    \textbf{81.5}\\
BiLSTM BERT-base & 75.4 & 87.6 & 72.6 & 80.8 & 79.1\\
BiLSTM SciBERT &  77.1 &    \textbf{89.9} &    72.1 &    \textbf{85.7} &    81.2\\
\midrule
BERT-base & 81.8 & 70.6 & 88.2 & 73.1 &  78.4\\
SciBERT & \textbf{84.5} & 77.0 & \textbf{91.6} & 72.7 & \textbf{81.5}\\
\midrule
\textit{humans} &  \textit{94.3} & \textit{95.9} & \textit{95.5} & \textit{97.5} & \textit{95.8}\\
\bottomrule
	\end{tabular}
\caption{\textbf{Experiments: entity mention detection and typing.} Results on test set (experiment-describing sentences only) in terms of F1, rightmost column shows the macro-average.}
\label{tab:results-entity-mention-extraction}
\end{table}
\paragraph{Experiment slot filling.}

\begin{table}[t]
	\centering
	\footnotesize
		\setlength{\tabcolsep}{10pt}
	\begin{tabular}{llr}
		\toprule
		\textbf{Model} & \textbf{dev} & \textbf{test}\\
		\midrule
		CRF & 45.3\tiny{+/-5.6} & 41.3\\
		\midrule
		BiLSTM mat2vec & 25.9\tiny{+/-11.2} & 22.5\\
		BiLSTM word2vec & 27.5\tiny{+/-9.0} & 27.0\\
		\hspace*{2mm}+ mat2vec & 43.0\tiny{+/-11.5} & 34.9\\
		\hspace*{4mm}+ bpe & 50.2\tiny{+/-11.8} & 38.9\\
		\hspace*{6mm}+ BERT-base & 64.6\tiny{+/-12.8} & 54.2\\
		\hspace*{6mm}+ SciBERT & 67.1\tiny{+/-13.3} & 59.7\\
		BiLSTM BERT-base & 63.3\tiny{+/-12.9} &   57.4\\
		BiLSTM SciBERT & \textbf{67.8}\tiny{+/-12.9} & \textbf{62.6}\\
		\midrule
		BERT-base & 63.4\tiny{+/-13.8} & 54.9\\
		SciBERT & 65.6\tiny{+/-13.2} & 56.4\\
		\midrule
		\textit{humans} & \multicolumn{2}{c}{\textit{83.4}}\\
		\bottomrule
	\end{tabular}
	\caption{\textbf{Experiments: slot identification}. Model comparison in terms of macro F1.}
	\label{tab:slotsMacro}
\end{table}

\tref{slotsMacro} shows the macro-average F1 scores for our different models on the slot identification task.\footnote{We evaluate on the 16 slot types as listed in \tref{slot-agreement}. When training our model, we use the additional types \textit{experiment\_evoking\_word} and \slotThickness, which are not frame slots but related annotations present in our data, see guidelines.}
As for entity typing, we train and evaluate our model on the subset of sentences marked as experiment-describing, which contain 4,263 slot instances.
Again, the CRF baseline outperforms the BiLSTM when using only mat2vec and/or word2vec embeddings.
The addition of BERT or SciBERT embeddings improves performance.
However, on this task, the BiLSTM model with (Sci)BERT embeddings outperforms the fine-tuned original (Sci)BERT model.
Compared to the other two tasks, this task requires more complex reasoning and has a larger number of possible output classes.
We assume that in such a setting, adding more abstraction power to the model (in the form of a BiLSTM) leads to better results.

For a more detailed analysis, \tref{slots} shows the slot-wise results for the non-neural CRF baseline and the model that performs best on the development set: BiLSTM with SciBERT embeddings.
As in the case of entity mention detection, the models do well for the categories that consist of numeric mentions plus particular units.
In general, model performance is also tied to the frequency of the slot types in the dataset.
Recognizing the role a material plays in an experiment (e.g., \slotAnodeMaterial vs. \slotCathodeMaterial) remains challenging, possibly requiring background domain knowledge.
This type of information is often not stated explicitly in the sentence, but introduced earlier in the discourse and would hence require document-level modeling.

\begin{table}[t]
	\centering
	\footnotesize
		\begin{tabular}{lrrrr}
		\toprule
		& & BiLSTM & \\
		& CRF & SciBERT & \textbf{count} \\
		\midrule
\slotAnodeMaterial & 25.0 & 19.0 & 280\\
\slotCathodeMaterial & 11.8 & 28.9 & 259\\
\slotDevice & 59.3 & 67.6 &381\\
\slotElectrolyteMaterial & 20.0 & 47.2 & 219\\
\slotFuelUsed & 45.9 & 55.5 & 159\\
\slotInterlayerMaterial & 0.0 & 10.7 & 51\\
\slotOpenCircuitVoltage & 43.5 & 84.3 & 44\\
\slotPowerDensity & 69.0 & 97.6 & 175\\
\slotResistance & 64.5 & 93.9 & 136\\
\slotWorkingTemperature & 72.5 & 90.3 & 414\\
\bottomrule
	\end{tabular}
	\caption{\textbf{Experiments: slot identification.} Results in terms of F1 on the test set, BiLSTM results averaged across 5 models.}
	\label{tab:slots}
\end{table}
\subsection{Entity Extraction Evaluation on the Synthesis Procedures Dataset}
\label{sec:synthesisExp}

\begin{table}[h]
	\centering
	\footnotesize
		\setlength{\tabcolsep}{3pt}
	\begin{tabular}{lr}
		\toprule
		\textbf{Model} & \textbf{micro-avg. F1}\\
		\midrule
		\textit{DCNN \cite{mysore2017automatically}} & 77.5\\
		\textit{BiLSTM-CRF \cite{mysore2017automatically}} & 77.6\\
		\midrule
		\midrule
		BiLSTM mat2vec & 73.9\\
		BiLSTM word2vec & 76.4 \\
		\hspace*{2mm}+ mat2vec & 83.5 \\
		\midrule
		BERT-base & 85.5 \\
		SciBERT & 87.2\\
		BiLSTM BERT-base & 89.3\\
		BiLSTM SciBERT & 90.7\\
		BiLSTM + all (with BERT-base) & 89.3\\
		BiLSTM + all (with SciBERT) & \textbf{92.2}\\
		\bottomrule
	\end{tabular}
	\caption{\textbf{Experiments: modeling mention types in synthesis procedure data set.} Results from \citet{mysore2017automatically} are not directly comparable to ours as they are based on a slightly different data set; our BiLSTM mat2vec+word2vec roughly corresponds to their BiLSTM-CRF model.}
	\label{tab:synthesisEntities1}
\end{table}

As described in \sref{relwork}, the data set curated by \citet{mysore2019materials} contains 230 synthesis procedures annotated with entity type information.\footnote{See \url{https://github.com/olivettigroup/annotated-materials-syntheses}}
We apply our models to this entity extraction task in order to estimate the degree of transferability of our findings to similar data sets.
To the best of our knowledge, there have not yet been any publications on the automatic modeling of this data set.
We hence compare to the previous work of \citet{mysore2017automatically}, who perform action graph induction on a similar data set.\footnote{According to correspondence with authors.}
Our implementation of BiLSTM-CRF mat2vec+word2vec roughly corresponds to their BiLSTM-CRF system.

\tref{synthesisEntities1} shows the performance of our models when trained and evaluated on the synthesis procedures dataset.
Detailed scores by entity type can be found in the Supplementary Material.
We chose to use the data split suggested by the authors for the NER task, using 200 documents for training, and 15 documents for each dev and test set.
Among the non-BERT-based systems, the BiLSTM variant using both mat2vec and word2vec performs best, indicating that the two pre-trained embeddings contain complementary information with regard to this task.
The best performance is reached by the BiLSTM model including word2vec, mat2vec, bpe and SciBERT embeddings, with 92.2 micro-average F1 providing a strong baseline for future work.

\section{Conclusion}
We have presented a new dataset for information extraction in the materials science domain consisting of \sofcCorpusSize open-access scientific articles related to solid oxide fuel cells.
Our detailed corpus and inter-annotator agreement studies highlight the complexity of the task and verify the high annotation quality.
Based on the annotated structures, we suggest three information extraction tasks: the detection of experiment-describing sentences, entity mention recognition and typing, and experiment slot filling.
We have presented various strong baselines for them, generally finding that BERT-based models outperform other model variants.
While some categories remain challenging, overall, our models show solid performance and thus prove that this type of data modeling is feasible and can lead to systems that are applicable in production settings.
Along with this paper, we make the annotation guidelines and the annotated data freely available.

\paragraph{Outlook.}
In \sref{synthesisExp}, we have shown that our findings generalize well by applying model architectures developed on our corpus to another dataset.
A natural next step is to combine the datasets in a multi-task setting to investigate to what extent models can profit from combining the information annotated in the respective datasets.
Further research will investigate the joint modeling of entity extraction, typing and experiment frame recognition.
In addition, there are also further natural language processing tasks that can be researched using our dataset.
They include the detection of events and sub-events when regarding the experiment-descriptions as events, and a more linguistically motivated evaluation of the frame-semantic approach to experiment descriptions in text, e.g., moving away from the one-experiment-per-sentence and one-sentence-per-experiment assumptions and modeling the graph-based structures as annotated.

\vspace*{2cm}

\section*{Acknowledgments}
We thank Jannik Str\"{o}tgen, Felix Hildebrand, Dragan Milchevski and everyone else involved in the Bosch MatKB project for their support of this research.
We also thank Stefan Gr\"{u}newald, Sherry Tan, and the anonymous reviewers for their insightful comments related to this paper.

\bibliography{references}
\bibliographystyle{acl_natbib}

\appendix

\renewcommand{\thesubsection}{\Alph{subsection}}
\section*{Supplementary Material}
\subsection{Background on Solid Oxide Fuel Cells}
\label{sec:supp:sofc}

A fuel cell is an electrochemical device that generates electricity exploiting the chemical reaction of a fuel (usually hydrogen) with an oxidant (usually air).
The reactions take place on two electrodes, the cathode and the anode, while the circuit is closed by an electrolyte material that only allows the transfer of charged atoms (see \fref{supp:sofc}).
Fuel cells that use a solid oxide as electrolyte (Solid Oxide Fuel Cells or SOFCs) are very efficient and cost-effective, but can only operate at high temperatures (500-1000°C), which can cause long start-up times and fast degradation.
SOFCs can be used as stationary stand-alone devices, to produce clean power for residential or industrial purposes, or integrated with other power generation systems to increase the overall efficiency.

\begin{figure}[h]
	\centering
\includegraphics[width=0.45\textwidth]{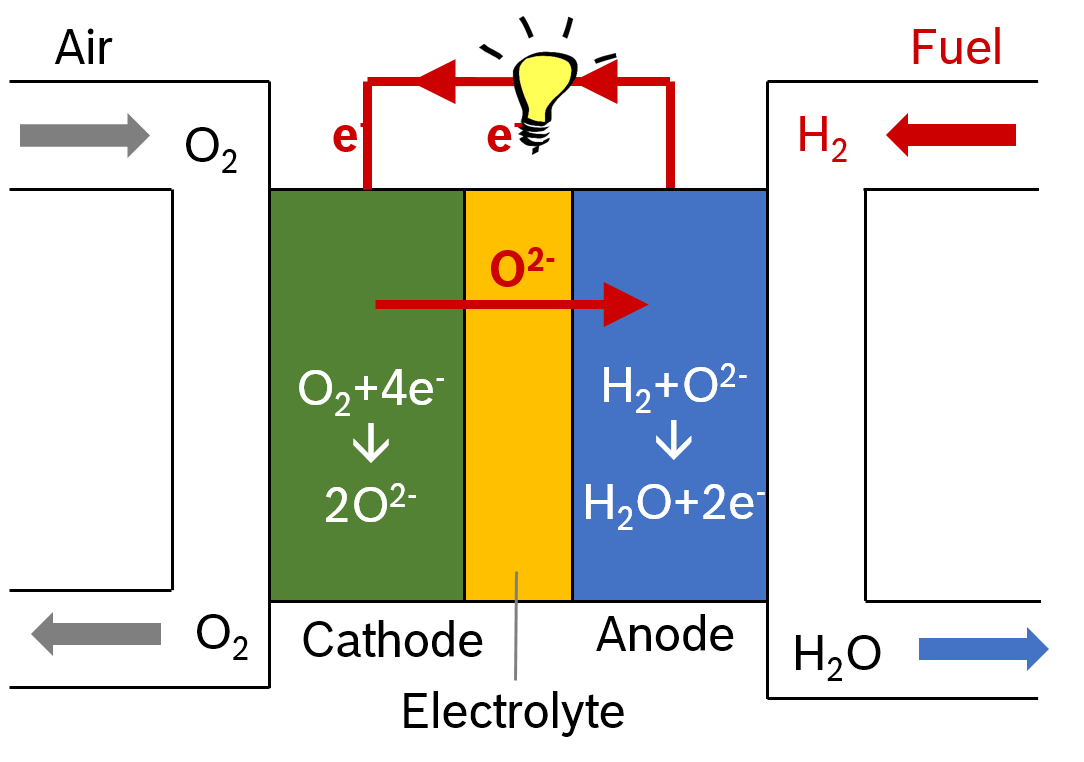}
\caption{Solid Oxide Fuel Cell schema.}
\label{fig:supp:sofc}
\end{figure}

\subsection{Data Analysis: Between-Experiment Links}
\label{sec:dataAnalysis}

As stated in \sref{annotationScheme}, we instructed annotators to mark the closest experiment-evoking word as \expLabelMention and link the respective slot arguments to this mention.
In addition, the \expLabelMention annotations could then be linked either by \sameExpLink or \varExpLink links.
\tref{between-experiment-links} shows some statistics on the number of \expLabelMention annotations per sentence and how often the primary annotator actually made use of the possibility to link experiments.
In the training data, out of 703 sentences describing experiments, 135 contain more than one experiment-evoking word, with 114 sentences containing two, 18 sentences containing three, and 3 sentences containing four \expLabelMention annotations (see \tref{between-experiment-links}).
In the 114	sentences containing two experiment annotations, only in 2 sentences, the \expLabelMention{}s were not linked to any others.
Upon being shown these cases, our primary annotator judged that one of them should actually have been linked.

Next, we analyze the number of cross-sentence links.
In the training data, there are 256 \sameExpLink and 93 \varExpLink links, of which 138 and 57 cross sentence-boundaries respectively.
Cross-sentence links between experiment-evoking words and slot fillers rarely occur in our dataset (only 13 out of 2,540 times).

\begin{table}
	\centering
	\footnotesize
	\begin{tabular}{lrrrr}
		\toprule
		\textbf{\# \expLabelMention} & \textbf{1} & \textbf{2} & \textbf{3} & \textbf{4}\\
		\textbf{per sentence} & \\
		\midrule
		\# sentences & 568 & 114 & 18 & 3\\
		\# \sameExpLink & 0 & 82 & 28 & 7\\
		\# \varExpLink &  0 & 27 & 8 & 1\\
		\midrule
		\# sent. with `unlinked' exp.& - & 2 & 1 & 0\\
		\bottomrule
	\end{tabular}
	\caption{\textbf{Data analysis.} Number of \expLabelMention annotations per sentence, and counts of links between them (within sentence). Training set: 703 experiment-describing sentences.}
	\label{tab:between-experiment-links}
\end{table}

\subsection{Inter-annotator Agrement Study: further statistics}
\label{sec:suppl:agreement}
\tref{slot-agreement} shows the full set of statistics for the experiment slot agreement.

\begin{table*}[h]
	\centering
	\footnotesize
	\begin{tabular}{lrrrrr}
		\toprule
		& \multicolumn{3}{c}{\textbf{agreement study}} & \textbf{IAA} &\textbf{train}\\
		& \multicolumn{1}{c}{\textbf{P}} & \multicolumn{1}{c}{\textbf{R}} & \multicolumn{1}{c}{\textbf{F1}} & \textbf{count} & \textbf{count}\\ \midrule
		\slotAnodeMaterial & 75.0 & 69.2 & 72.0 & 13 & 280\\
		\slotCathodeMaterial & 84.8 & 88.6 & 86.7 & 44 & 259\\
		\slotConductivity & - & - & - & - & 55\\
		\slotCurrentDensity & 100.0 & 60.0 & 75.0 & 5 & 65\\
		\slotDegradationRate & 100.0 & 100.0 & 100.0 & 2 & 19\\
		\slotDevice & 97.1 & 93.0 & 95.0 & 71 & 381\\
		\slotElectrolyteMaterial & 78.9 & 93.8 & 85.7 & 48 & 219\\
		\slotFuelUsed & 90.0 & 81.8 & 85.7 & 11 & 159\\
		\slotInterlayerMaterial & 100.0 & 56.0 & 71.8 & 25 & 51\\
		\slotOpenCircuitVoltage & 90.0 & 90.0 & 90.0 & 10 & 44\\
		\slotPowerDensity & 100.0 & 85.1 & 92.0 & 47 & 175\\
		\slotResistance & 100.0 & 100.0 & 100.0 & 26 & 136\\
		\slotSupportMaterial & 75.0 & 37.5 & 50.0 & 8 & 106\\
		\slotTimeOfOperation & 83.3 & 100.0 & 90.9 & 5 & 47\\
		\slotVoltage & 100.0 & 33.3 & 50.0 & 6 & 35\\
		\slotWorkingTemperature & 98.6 & 94.5 & 96.5 & 73 & 414\\
		\bottomrule
	\end{tabular}
	\caption{\textbf{Inter-annotator agreement study.} Precision, recall and F1 scores of the two annotators vs. each other on the set of \textbf{slots}. \textbf{IAA count} refers to the number of mentions labeled with the respective type by our primary annotator in the 6 documents of the inter-annotator agreement study. \textbf{train count} refers to the number of instances in the training set. (\slotConductivity has been added to the set of slots only after conducting the inter-annotator agreement study.)}
	\label{tab:slot-agreement}
\end{table*}

\subsection{Additional Experimental Results}
\label{sec:suppl-scores}

In the following tables, we give detailed statistics for the experiments described in the main paper.
\begin{description}
	\item[\tref{results-experiment-detection_sup}] reports full statistics for the task of identifying experiment-describing sentences, including precision and recall in the dev setting.
	\item[\tref{results-entity-mention-extraction-sup}] reports F1 per entity type for the dev setting including standard deviations.
	\item[\tref{synthesisEntities2}] reports F1 per entity type/slot for the synthesis procedures dataset \citep{mysore2019materials}.
\end{description}

\begin{table*}[h]
	\centering
	\footnotesize
	\begin{tabular}{lrrr|rrr|}
		\toprule
		& \multicolumn{3}{c|}{\textbf{dev (5-fold cv)}}& \multicolumn{3}{c|}{\textbf{test}}\\
		Model & P & R & F1 & P & R & F1\\
		\midrule
		RBF SVM &  66.4 &    46.1 &    54.2\tiny{+/-3.7} &    64.6 &    54.9 &    59.4\\
		Logistic Regression & \textbf{72.7} &    41.9 &    53.0\tiny{+/-4.2} &    \textbf{68.2} &    50.9 &    58.3\\
		\midrule
		BiLSTM mat2vec &    46.3 &    55.6 &    49.9\tiny{+/-3.1} &    49.6 &    69.4 &    57.8\\
		BiLSTM word2vec &    50.0 &    56.1 &    52.3\tiny{+/-4.6} &    51.1 &    65.3 &    57.4\\
		\hspace*{2mm}+ mat2vec &    59.8 &    53.6 &    55.9\tiny{+/-4.2} &    52.0 &    59.0 &    55.3\\
		\hspace*{4mm}+ bpe &    62.2 &    56.4 &    58.6\tiny{+/-3.0} &    58.9 &    64.7 &    61.7\\
		\hspace*{6mm}+ BERT &    66.1 &    67.8 &    66.8\tiny{+/-4.9} &    60.2 &    71.7 &    65.4\\
		\hspace*{6mm}+SciBERT & 68.6 &    68.0 &    \textbf{68.1}\tiny{+/-3.7} &    60.2 &    73.4 &    66.1\\
		BiLSTM BERT &    65.5 &    64.2 &    64.7\tiny{+/-4.6} &    63.7 &    69.9 &    66.7\\
		BiLSTM SciBERT &    67.1 &    69.1 &    67.9\tiny{+/-4.0} &    58.6 &    74.6 &    65.6\\
		\midrule
			BERT-base & 64.0 &    68.2 & 66.0\tiny{+/-4.6} &    58.6 &   71.1 &  64.2\\
			BERT-large &    61.8 &    68.9 &    64.3\tiny{+/-4.6} &    63.1 &    \textbf{75.1} &    \textbf{68.6}\\
			SciBERT &    66.0 &    \textbf{70.2} &    67.9\tiny{+/-4.0} &    60.8 &    74.6 &    67.0\\
		\midrule
		\textit{humans (on agreement data)} & \textit{80.4} & \textit{77.6} & \textit{78.9} & \textit{80.4} & \textit{77.6} & \textit{78.9}\\
		\bottomrule
	\end{tabular}
	\caption{\textbf{Experiments: Identifying experiment sentences.} P, R and F1 for experiment-describing sentences. With the exception of SVM, we downsample the non-experiment-describing sentences by 0.3.}
	\label{tab:results-experiment-detection_sup}
\end{table*}

\begin{table*}[h]
	\centering
	\footnotesize
	\hspace*{-3mm}
	\begin{tabular}{lrrrrp{1.2cm}|rrrrp{1.2cm}}
		\toprule
		\textbf{Model} & \rot{\expLabelMention} & \rot{\matLabel} & \rot{\valLabel} &
		\rot{\devLabel} & \rot{\textbf{macro-avg.}} & \rot{\expLabelMention} & \rot{\matLabel} & \rot{\valLabel} &
		\rot{\devLabel} & \rot{\textbf{macro-avg.}}\\
		\midrule
		CRF & 66.5\tiny{+/-3.5} & 47.0\tiny{+/-9.1} & 73.0\tiny{+/-6.4} & 56.2\tiny{+/-10.0} & 60.7\tiny{+/-4.5} 
		& 61.4 & 42.3 & 73.6 & 64.1 & 60.3\\
		\midrule
		BiLSTM mat2vec &    52.9\tiny{+/-3.4} &    55.3\tiny{+/-2.0} &    47.9\tiny{+/-6.3} &    53.2\tiny{+/-1.9} &    52.3\tiny{+/-3.4} &    47.1 &    52.4 &    60.9 &    46.1 &    51.6\\
		+ BERT & 80.3\tiny{+/-3.2} &    87.7\tiny{+/-3.3} &    76.8\tiny{+/-5.3} &    81.9\tiny{+/-5.5} &    \textbf{81.7}\tiny{+/-4.3} &    74.3 &    87.9 &    71.0 &    80.7 &    78.5\\
		BiLSTM word2vec &    62.3\tiny{+/-3.0} &    61.6\tiny{+/-2.1} &    52.1\tiny{+/-5.2} &    59.5\tiny{+/-1.0} &    58.9\tiny{+/-2.8} &    55.8 &    58.6 &    59.1 &    51.7 &    56.3\\
		\hspace*{2mm}+mat2vec &    65.8\tiny{+/-4.2} &    78.4\tiny{+/-1.6} &    61.9\tiny{+/-8.2} &    69.6\tiny{+/-4.0} &    68.9\tiny{+/-4.5} &    57.9 &    75.2 &    64.3 &    61.5 &    64.7\\
		\hspace*{4mm}+bpe &    69.2\tiny{+/-5.8} &    82.3\tiny{+/-1.9} &    60.1\tiny{+/-11.2} &    73.4\tiny{+/-4.7} &    71.2\tiny{+/-5.9} &    63.3 &    81.6 &    68.0 &    68.1 &    70.2\\
		\hspace*{6mm}+BERT &    80.0\tiny{+/-3.4} &    87.9\tiny{+/-2.8} &    74.4\tiny{+/-5.6} &    80.7\tiny{+/-3.9} &    80.8\tiny{+/-3.9} &    76.0 &    88.1 &    72.9 &    81.5 &    79.7\\
		\hspace*{6mm}+SciBERT & 81.4\tiny{+/-1.6} &    \textbf{89.4}\tiny{+/-2.4} &    73.8\tiny{+/-8.7} &    82.0\tiny{+/-4.3} &    \textbf{81.7}\tiny{+/-4.2} &    76.9 &    89.8 &    74.1 &    85.2 &    \textbf{81.5}\\
		BiLSTM BERT & 79.6\tiny{+/-2.4} &    87.6\tiny{+/-2.4} &    72.0\tiny{+/-7.5} &    80.5\tiny{+/-5.1} &    79.9\tiny{+/-4.3} &    75.4 &    87.6 &    72.6 &    80.8 &    79.1\\
		BiLSTM SciBERT & 80.5\tiny{+/-1.2} &    \textbf{89.4}\tiny{+/-2.8} &    73.0\tiny{+/-9.4} &    \textbf{82.3}\tiny{+/-3.5} &    81.3\tiny{+/-4.2} &    77.1 &    \textbf{89.9} &    72.1 &    \textbf{85.7} &    81.2\\
		\midrule
			BERT-base &    \textbf{85.4}\tiny{+/-2.8} &    73.7\tiny{+/-7.2} &   90.0\tiny{+/-2.1} &  68.3\tiny{+/-3.7} &  79.3\tiny{+/-3.9} &    \textbf{81.8} &    70.6 &    88.2 &  73.1 &    78.4\\
			SciBERT &  84.5\tiny{+/-3.0} &    77.0\tiny{+/-7.4} &    \textbf{91.6}\tiny{+/-2.8} &    72.7\tiny{+/-2.1} &    81.5\tiny{+/-3.8} &    81.2 &    75.3 &    \textbf{91.9} &    73.2 &    80.4\\
		\midrule
		\textit{humans} & \textit{94.3} & \textit{95.9} & \textit{95.5} & \textit{97.5} & \textit{95.8}  & \textit{94.3} & \textit{95.9} & \textit{95.5} & \textit{97.5} & \textit{95.8}\\
		\bottomrule
	\end{tabular}
	\caption{\textbf{Experiments: entity mention extraction and labeling.} Results on 5-fold cross validation for dev and test set (experiment-describing sentences only) in terms of F1.}
	\label{tab:results-entity-mention-extraction-sup}
\end{table*}

\begin{table*}[h]
	\centering
	\footnotesize
	\setlength{\tabcolsep}{10pt}
	\begin{tabular}{lrrr}
		\toprule
		\textbf{Entity Types} & Mysore et & BiLSTM & BiLSTM\\
		& al. (2017) & w2v+m2v & + all (SciBERT)\\
		\midrule
		Amount-Unit & 83.5 & 93.5&  \textbf{95.8} \\
		Brand & - & 67.9 & \textbf{83.3} \\
		Condition-Misc & 74.6  & 85.1 & \textbf{88.9}\\
		Condition-Unit & 94.5 & \textbf{97.2} & 95.0 \\
		Material & 80.2  & 84.0 & \textbf{92.3}\\
		Material-Descriptor* & 62.0 & 65.5 & \textbf{88.5}  \\
		Nonrecipe-Material & - & 45.8 & \textbf{80.0} \\ 
		Number & 91.9  & 93.4 & \textbf{98.4} \\
		Operation & 82.8  & 93.5 & \textbf{98.1}\\
		Synthesis-Apparatus & - & 63.9 & \textbf{81.3}  \\
		\bottomrule
	\end{tabular}
	\caption{\textbf{Experiments: Modeling mention types in synthesis procedure data, most frequent entity types.} Results in terms of F1.
		Results from \citet{mysore2017automatically} are not directly comparable. 	*Type called Descriptor in their paper.}
	\label{tab:synthesisEntities2}
\end{table*}

\end{document}